\documentclass{article}

\usepackage[preprint]{neurips_2022}

\usepackage{amsmath,amsfonts,bm}

\def\eqref#1{equation~\ref{#1}}
\def\1{\bm{1}}

\def\vdelta{{\bm{\delta}}}

\def\vpsi{{\bm{\psi}}}

\def\vx{{\bm{x}}}
\def\vy{{\bm{y}}}

\DeclareMathAlphabet{\mathsfit}{\encodingdefault}{\sfdefault}{m}{sl}
\SetMathAlphabet{\mathsfit}{bold}{\encodingdefault}{\sfdefault}{bx}{n}

\def\sR{{\mathbb{R}}}

\DeclareMathOperator*{\argmax}{arg\,max}

\usepackage[utf8]{inputenc} % allow utf-8 input
\usepackage[T1]{fontenc}    % use 8-bit T1 fonts
\usepackage{hyperref}       % hyperlinks
\usepackage{url}            % simple URL typesetting
\usepackage{booktabs}       % professional-quality tables
\usepackage{amsfonts}       % blackboard math symbols
\usepackage{nicefrac}       % compact symbols for 1/2, etc.
\usepackage{microtype}      % microtypography
\usepackage{xcolor}         % colors
\usepackage{graphicx}
\usepackage[capitalize]{cleveref}
\usepackage{algorithm}
\usepackage[noend]{algpseudocode}
\usepackage{adjustbox}

\usepackage[skip=11pt]{caption}
 
\usepackage{enumitem}
  \newlist{inlinelist}{enumerate*}{1}
  \setlist*[inlinelist,1]{%
          label=(\roman*),
      }

\newcommand{\asp}{\textsc{ASP}}

\newcommand{\miniimagenet}{\textit{mini}ImageNet}
\newcommand{\cnaps}{\textsc{CNAPs}}
\newcommand{\protonets}{ProtoNets}
\newcommand{\metadataset}{\textsc{Meta-Dataset}}

\title{Adversarial Attacks are a Surprisingly Strong Baseline for Poisoning Few-Shot Meta-Learners}

\author{%
  Elre T.~Oldewage%\thanks{Use footnote for providing further information
  \\
  University of Cambridge\\
  \texttt{etv21@cam.ac.uk} \\
  \And
  John Bronskill\\
  University of Cambridge\\
  \texttt{jfb54@cam.ac.uk} \\
  \And
  Richard E.~Turner\\
  University of Cambridge\\
  \texttt{ret26@cam.ac.uk} \\
}

\begin{document}

\maketitle

\begin{abstract}
This paper examines the robustness of deployed few-shot meta-learning systems when they are fed an imperceptibly perturbed few-shot dataset. 
We attack amortized meta-learners, which allows us to craft colluding sets of inputs that are tailored to fool the system's learning algorithm when used as training data.
Jointly crafted adversarial inputs might be expected to synergistically manipulate a classifier, allowing for very strong data-poisoning attacks that would be hard to detect.
We show that in a white box setting, these attacks are very successful and can cause the target model's predictions to become worse than chance.
However, in opposition to the well-known transferability of adversarial examples in general, the colluding sets do not transfer well to different classifiers.
We explore two hypotheses to explain this: ``overfitting'' by the attack, and mismatch between the model on which the attack is generated and that to which the attack is transferred.
Regardless of the mitigation strategies suggested by these hypotheses, the colluding inputs transfer no better than adversarial inputs that are generated independently in the usual way. 
\end{abstract}
%We show that inserting adversarial examples into the support set of a meta-learner has a significant effect on resulting predictions.  the resulting predictions on test inputs can become worse than chance, the .
%
%This is achieved by developing a novel attack, \textit{Adversarial Support Poisoning} or \asp{}, which crafts a poisoned \textit{set} of examples.
%
%When even a small subset of malicious data points is inserted into the support set of a meta-learner, accuracy is significantly reduced.
%
%We evaluate the new attack on a variety of few-shot classification algorithms and scenarios, and propose a form of adversarial training that significantly improves robustness against both poisoning and evasion attacks.

\section{Introduction}
\label{sec:introduction}
Standard deep learning approaches suffer from poor sample efficiency \citep{krizhevsky2012imagenet} which is problematic in tasks where data collection is difficult or expensive. Few-shot learners have been developed to address this shortcoming by supporting rapid adaptation to a new task using only a few labeled examples \citep{finn2017model, snell2017prototypical}.
This success has made few-shot learners more attractive for increasingly sensitive applications where the repercussions of confidently-wrong predictions are severe, such as clinical risk assessment \citep{xi2019metapred}, glaucoma diagnosis \citep{kim2017glaucoma} and diseases identification in skin lesions \citep{mahajan2020metadermdiagnosis}.
%, and tissue slide annotation in cancer immuno-therapy biomarker research \citep{lahiani2018immunohistochemistry}.

As few-shot learners gain popularity, it is essential to understand how robust they are and how they may be exploited. It is well known that standard classifiers are vulnerable to \textit{adversarial} inputs which have been purposefully and imperceptibly modified to cause incorrect predictions \citep{biggio2017wildpatterns}. Such examples may be presented to a model either at test time, called \textit{evasion attacks} \citep{biggio2017evasion} or \textit{adversarial examples} \citep{Szegedy14intriguingproperties}, or at training time, which is referred to as \textit{poisoning} \citep{newsome2006thwarting,rubinstein2009antidote}. 
The concepts of evasion and poisoning can be re-stated in terms of few-shot learners. At meta-test time, a few-shot learner is presented with an unseen task containing a few labeled examples, the \textit{support set}, and a number of unlabeled examples to classify, called the \textit{query set}. We can perpetrate a ``standard'' adversarial attack by perturbing an image in the query set or perform a poisoning attack by perturbing images in the support set. 

In this paper, we propose a poisoning attack against amortized meta-learners that \textit{jointly} optimizes attack images \textit{through the meta-learner's adaptation mechanism}. We produce a colluding \textit{set} of adversarial images that is explicitly tailored to fool the learning algorithm. 
We consider as a baseline the insertion of standard adversarial attack images into the support set. We expect our attack to perform better than this baseline, since standard adversarial attacks are optimised \textit{individually} to cause misclassification by a fixed classifier (rather than being curated as a set to fool the adaptation mechanism itself). 
We may also expect the new attack to transfer to other learners, especially considering the well-known transferability of adversarial examples in literature \citet{goodfellow2014explaining,papernot2016transferability,Szegedy14intriguingproperties}.
%
%While previous work has considered evasion attacks, in the context of few shot learners \citep{goldblum2019adversarially, yin2018adversarial}, data poisoning attacks have not been studied and are the focus of this paper. 
Our contributions are as follows:
\begin{enumerate}[noitemsep,nolistsep]
\item We define a poisoning attack on few-shot classifiers, referred to as \textbf{adversarial support poisoning} (ASP) or simply as \textit{support attacks}. This applies coordinated adversarial perturbations to produce a colluding set of inputs, generated by jointly backpropagating through the meta-learner's adaptation process to minimize model accuracy over a set of query points. To the best of the authors' knowledge, ours is the first work to perpetrate poisoning attacks on trained few-shot classifiers. 
\item In a white box setting \asp{} is a very strong attack, far more effective than the baseline.
\item Surprisingly, we show that \asp{} attacks do not transfer significantly better than the \textbf{swap attack} baseline, which inserts adversarial examples generated ``in the usual way'' into the support set.
\item We observe that transferability of support attacks are highly dependent on the feature extractor used by the meta-learner. Although we propose various mitigation strategies to improve the transferability of support set attacks, we do not succeed at significantly improving over the efficacy of a swap attack, which is simpler and cheaper.
\end{enumerate}
The last two findings are unexpected: we expect the \asp{} attack to be significantly stronger than the swap attack baseline because it is tailored to fool the learner's adaptation process, is optimized over a set of query points and is able to collude within the adversarial set.
% i.e. having been generated by backpropagation through the whole adaptation process, not just the adapted metalearner.
%We proceed as follows: Section~\ref{sec:background} provides the necessary background, \cref{sec:attacking_learners} introduces the ASP attack, \cref{sec:experiments} presents the experimental results and \cref{sec:conclusion} concludes the paper.
%
%We consider attacking a few-shot meta-learner at meta-test time, i.e. by inserting perturbed examples into its support set to cause changes in predictions on its target set.
%
%We show that simply inserting classical adversarial imamges into the support set is a very effective way to reduce a few-shot classifier's accuracy and that this transfers about as well as the more expensive poisoning attack.

\section{Background}
\label{sec:background}
We focus on image classification, denoting input images by $x \in \sR^{ch \times W \times H}$ where $W$ is the image width, $H$ the image height, $ch$ the number of image channels and image labels $y \in \{1,\hdots, C\}$ where $C$ is the number of image classes. We use bold $\vx$ and $\vy$ to denote sets of images and labels. 
We consider the few-shot image classification scenario using a  meta-learning approach.
Rather than a single, large dataset $D$, we assume access to a dataset $\mathcal{D} = \{ \tau_t \}_{t = 1}^{K}$ comprising a large number of training \textit{tasks} $\tau_t$, drawn i.i.d.~from a distribution $p(\tau)$. 
An example task is shown in \cref{fig:task}. 
The data for a task consists of a \textit{support set} $D_S=\{(x_n, y_n)\}_{n=1}^{N}$ comprising $N$ elements, with the inputs $x_n$ and labels $y_n$ observed, and a \textit{query set} $D_Q=\{(x^{\ast}_m, y^{\ast}_m)\}_{m=1}^{M}$ with $M$ elements for which we wish to make predictions. We may use the shorthand $D_S=\{\vx, \vy\}$ and $D_Q=\{\vx^{\ast}, \vy^{\ast}\}$.
The meta-learner $g$ takes as input the support set $D_S$ and produces task-specific classifier parameters $\vpsi = g(D_S)$ which are used to adapt the classifier $f$ to the current task. The classifier can now make task-specific predictions $f(x^{\ast}, \vpsi= g(D_S))$ for any test input $x^{\ast} \in D_Q$.
Here the inputs $x^{\ast}$ are observed and the labels $y^{\ast}$ are only observed during meta-training (i.e.~training of the meta-learning algorithm).
Note that the query set examples are drawn from the same set of labels as the examples in the support set.
The majority of modern meta-learning methods employ \textit{episodic} training \citep{vinyals2016matching}, as detailed in \ref{app:episodic_training}.
At meta-test time, the classifier $f$ is required to make predictions for query set inputs of unseen tasks, which will often include classes that have not been seen during meta-training, and $D_S$ will contain only a few observations.

The canonical example for modern gradient-based few-shot learning systems is MAML \citep{finn2017model}. Another widely used class of meta-learners are \textit{amortized-inference} or \textit{black box} based approaches e.g, \textsc{Versa} \citep{gordon2018meta} and \textsc{CNAPs} \citep{requeima2019cnaps}. 
In these methods, the task-specific parameters $\vpsi$ are generated by one or more \textit{hyper-networks}, $g$ \citep{ha2016hypernetworks}.
Prototypical Networks (\protonets{}) \citep{snell2017prototypical} is a special case of this which is based on \textit{metric} learning and employs a nearest neighbor classifier.
We focuses on Simple \cnaps{} \cite{Bateni2020improved} -- which improves on the performance of \cnaps{} -- and uses a \protonets{} classifier head with an adaptive feature extractor like that of \cnaps{}; but we have also considered attacks against MAML, \protonets{}, and \cnaps{} with similar outcomes.

\section{Attacking Few-Shot Learners}
\label{sec:attacking_learners}

We summarize the threat model in terms of the adversary's goal, capabilities and knowledge. In this work, we develop poisoning attacks that degrade the model \textit{availability} (i.e.~affect prediction results indiscriminately such that it is useless) \citep{jagielski2018manipulating}. We allow the attacker to manipulate some fraction of the support set and further constrain pattern modifications to be imperceptible (i.e. within some $\epsilon$ of the original image, measured using the $\ell_\infty$ norm). 
When considering transfer attacks, we call the model used generate the attack the \textit{surrogate} and the model to which we transfer the attack the \textit{target}. We presume access to the surrogate model's gradients and internal state. We don't assume any access to the target model's internal state, though some of the experiments do access a limited number of the target's predictions. We also assume access to enough data to form a query set. 

Consider a target model that has been trained and deployed as a service. As described in Section~\ref{sec:introduction}, a malicious party could perpetrate two kinds of attacks: a ``standard'' adversarial attack (hereafter referred to as a \textit{query} attack) by perturbing elements in the query set of a task; or perform a poisoning attack by perturbing elements in the support set. We expand on these below:
\paragraph{Query Attack} Given a meta-learner that has already adapted to a specific task, the attacker perturbs a \textit{single} test input $x^{\ast}$ in such a way that \textit{it} will be misclassified. 
This corresponds to solving %
$\argmax_{{\delta}} \mathcal{L}(f(x^{\ast} + \delta, g(\vx, \vy)),y^{\ast})$. 
%
%In this situation, the goal is for the \textit{specific} input $x^{\ast}$ to be misclassified. 
Refer to \cref{app:pgd_query_algorithm} for details. % Here we change a test input
These attacks are essentially evasion attacks as considered in \citet{biggio2017evasion} and many algorithms can be used to generate adversarial examples \citep{madry2017towards, carlini2017towardsevaluating, chen2017EAD}.
Query attacks have been perpetrated successfully against few-shot learners \citep{goldblum2019adversarially, yin2018adversarial}.
%
%
%Here, we change the dataset
\paragraph{Adversarial Support Poisoning (ASP)} The attacker perturbs the \textit{dataset} that the target meta-learner will learn from so as to influence \textit{all future test predictions}.
The attacker thus computes a perturbed support set $\tilde{D}_S = \{ \tilde{\vx}, \vy\} $ whose inputs are jointly optimized to fool the system on a specific query set, which we call the \textit{seed query set}, with the goal of generalizing to unseen query sets.
%The attacker may want the system to fail on \textit{any} query image. 
This corresponds to solving 
$\argmax_{\vdelta} \mathcal{L}(f(\vx^{\ast}, g(\vx + \boldsymbol{\delta}, \vy)),\vy^{\ast})$ such that $\|\boldsymbol{\delta}\|_\infty < \epsilon$, where $D_Q =  \{\vx^{\ast}, \vy^{\ast}\}$ denotes the seed query set and $\epsilon$ is the maximum size of the perturbation.
Refer to \cref{alg:pgd} for details.
%
%We call this novel few-shot learner attack \textit{Adversarial Support Poisoning}, or simply \asp{}. 
%
Our attack is a poisoning attack, since the attacker is manipulating data that the model will use to do inference. However, it is important to note the the attack is perpetrated at meta-test time, after the meta-learner has already been meta-trained.
%
%Example images from an \asp{} attack are shown in \cref{app:perturbed_examples}.
%
Unlike a query attack, which is generated by backpropagating through the adapted learner with respect to a single point, our attack backpropagates jointly through the adaptation process with respect to the loss on the entire query set and should thus be a much stronger poisoning attack than say, inserting a query point into the support set.
%
%In real settings, an attacker might design an attack on their own query set, hoping it will generalize to unseen queries. The ability to generalize may depend on $M$, the size of the seed query set. 
Without loss of generality, we use Projected Gradient Descent (PGD) \citep{madry2017towards} to generate perturbed support sets because it is effective, simple to implement, and easily extensible to sets of inputs.

\begin{algorithm}[h]
\caption{PGD for \asp{}}
\label{alg:pgd}
%\begin{multicols}{2}
\begin{algorithmic}[1]
\Require
\Statex $I_{min}, I_{max}$: Min/Max image intensity, $\gamma$: Step size, $L$: Number of iterations, 
\Statex $D_S \equiv \{\vx,\vy\}, D_Q \equiv \{\vx^*,\vy^*\}$, $\mathcal{L} \equiv$ cross-entropy, $\epsilon$: Perturbation amount
\Procedure{PGDS}{$D_S,D_Q,f,g$}
\State $\vdelta \sim U(-\epsilon, \epsilon)$
\State $\tilde \vx \leftarrow \text{clip}(\vx + \vdelta, I_{min}, I_{max})$ 
\For{$i \in 1,...,L$}
\State $\vdelta \leftarrow \text{sgn}(\nabla_{\tilde \vx}\mathcal{L}(f(\vx^{\ast}, g(\tilde \vx, \vy)),\vy^{\ast}))$
\State $\tilde \vx \leftarrow \text{clip}(\tilde \vx + \gamma\vdelta, I_{min}, I_{max})$
\State $\tilde \vx \leftarrow \vx + \text{clip}(\tilde \vx - \vx, -\epsilon, \epsilon)$
\EndFor
\State \Return $\tilde \vx $
\EndProcedure
\end{algorithmic}
\end{algorithm}

\section{Experiments}
\label{sec:experiments}
%In this section we present our experiments that endeavor to answer the following questions:
%\begin{inlinelist}
%\item How vulnerable are few-shot classifiers to \asp{} attacks?
%\item What are the most effective parameter settings for \asp{}?
%\item Is adversarial training effective at mitigating \asp{} attacks?
%\end{inlinelist}
%
The experiments presented in the main body of the paper are carried out on Simple \cnaps{} \citep{Bateni2020improved} using the challenging \metadataset{} benchmark (refer to  \cref{app:training_protocols} and for details on the meta-data training protocols and \metadataset{}). The Simple \cnaps{} model uses a \protonets{} head in conjunction with an adaptive feature extractor. The \protonets{} head uses Euclidean distance and the feature extractor is a Resnet18 endowed with Feature-wise Linear Modulation (FiLM) layers \citep{perez2018FiLM}, which scales and shifts a feature map $\mathbf{x}$ by $\zeta \mathbf{x} + \beta$. In the meta-learning setting, FiLM parameters are generated by an adaptation network as part of $\vpsi$, the learner's adaptation to a new task, and allows the feature extractor to adapt flexibly to each new task with relatively few additional parameters \citep{requeima2019cnaps}.
In \metadataset{}, task support sets may be large --- up to 500 images across all classes. In a realistic scenario, an attacker would not likely be able to perturb all the images in such a large support set, so we only perturb a specified fraction of the support set in each experiment.
%We thus perturb only 20$\%$ of the images in the support set, which is generally considered the upper limit of manipulated patterns in conventional poisoning approaches \citep{jagielski2018manipulating}.  
%
%For all experiments, we consider the classifier's performance averaged over 500 randomly generated tasks. 
Each task is composed of a support set, a seed query set and up to 50 unseen query sets used for attack evaluation (some \metadataset{} benchmarks do not have sufficiently many patterns available to form 50 query sets and thus were excluded from the experiments). The unseen query sets are all disjoint from the seed query set to avoid information leakage. 
For each task, we generate an adversarial support set using the original support set and corresponding seed query set. The adversarial support set is then evaluated on the task's unseen query sets. 
We refer to the average classification accuracy on the seed query sets as the \textit{\asp{} Specific} attack accuracy, and when evaluating the attack on unseen query sets we refer to it as the \textit{\asp{} General} attack accuracy. 
An important baseline is the \textbf{swap attack}, which is generated by using the task's support set and query set to produce a query attack, then ``swap'' the role of the adversarial query set by using it as a support set. Like our \asp{} attack, the swap attack is evaluated on unseen query sets. Note that the swap attack is far less expensive to compute than the \asp{} attack, since it only requires backpropagation through the adapted learner and not through the entire meta-learning process. %However, it proves to be a surprisingly strong baseline when considering transfer attacks.

\subsection{White Box Attack}
\label{sec:white_box}

\begin{figure*}
	\centering
	\includegraphics[width=0.9\textwidth]{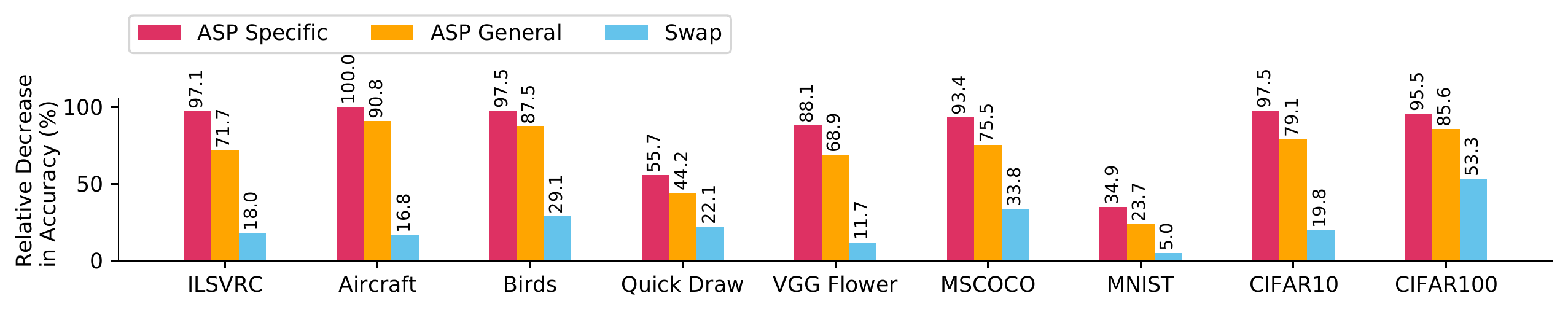}
	\caption{Relative drop in model accuracy attacking Simple \cnaps{} on \metadataset{} with $\epsilon=0.05$, $\gamma=0.0015$, $L = 100$, with all classes, but only $20\%$ of the shots poisoned.}
	\label{fig:basic_meta_dataset}
\end{figure*}
We first consider the simplest scenario, a white box attack in which the surrogate and target model are the same. \cref{fig:basic_meta_dataset} shows the relative decrease in accuracy of Simple \cnaps{} on \metadataset{}. %
In general, the clean accuracy of the surrogate and target models may be quite different and so we normalise the effect an attack has by computing the percentage relative decrease in classification accuracy as follows: $100\%\times(a_{clean} - a_{attack})/a_{clean}$ where $a_{clean}$ is the \textit{clean} classification accuracy before the attack, and $a_{attack}$ is the classification accuracy after the attack. Unnormalised results are in \ref{app:large_scale_results}.
Although it is expected for \textit{\asp{} Specific} to be very effective, 
%is highly effective, which is expected, since the attack is being evaluated on the query set used to generate the attack.
%on all datasets except MNIST, which is the easiest classification problem in the large-scale suite because there are only ten classes and the input images are simplistic.
%
the \asp{} attack remains highly effective even on unseen query sets (see \textit{\asp{} General}), easily out-performing the \textit{Swap} baseline, in spite of the fact that only $20\%$ of the support set shots are poisoned. At least in the white box setting, \asp{} is very strong attack.
%Our results demonstrate that an attacker using \asp{} could cripple a few-shot learning system in this white box scenario -- with a challenging dataset and a limited number of perturbed patterns -- far more effectively than with a simple swap attack.

\subsection{Attack Transfer}
\label{sec:transference}
\begin{figure*}
	\centering
	\includegraphics[width=\textwidth]{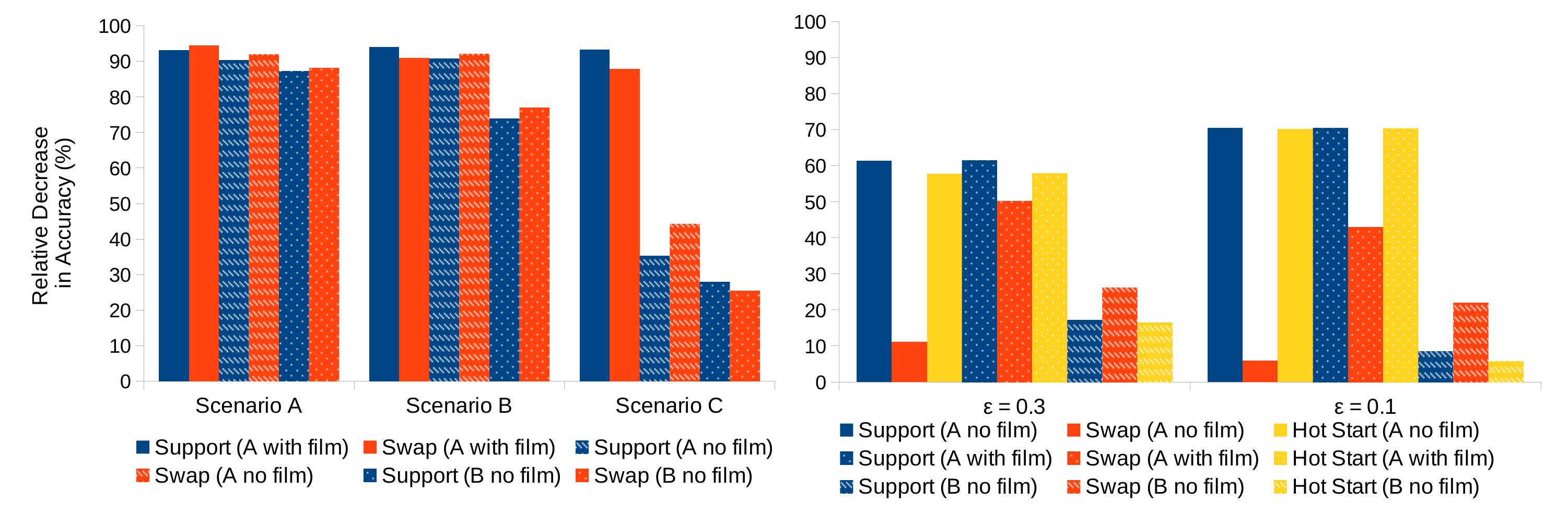}
	\caption{Transfer to Simple \cnaps{} with different backbones. Attacks on the surrogate model are solid fill; transfer attacks have diagonal hatching for Resnset18 A with FiLM and dots for Resnet18 B no FiLM. \textbf{(Left)} Scenario A: $\epsilon = 1.0$ with entire support set adversarial; Scenario B: $\epsilon=0.3$ with entire support set adversarial; Scenario C: $\epsilon=0.3$ with 50\% of support set adversarial. Dataset is CIFAR-100; $L=100$. \textbf{(Right)} \asp{}, hot-started \asp{} and swap attacks on ILSVRC 2012; $L=50$.}  
	\label{fig:protonets_easy_backbones}
\end{figure*}

We now turn our attention to the more difficult scenario, in which the target and surrogate models are different. We fix our surrogate model as Simple \cnaps{} with a particular Resnet18 feature extractor (hereafter referred to Resnet18 A). 
%Ideally, we may want to consider a target model with a completely different feature extractor and learning mechanism. But in the first instance, 
We consider a Simple \cnaps{} model with a slightly different feature extractor: Resnet18 B, which was trained on the same data with similar training conditions, but with no FiLM parameters. We also consider turning the FiLM parameters for Resnet18 A on and off. Although this seems a rather trivial extension, the results shown in \cref{fig:protonets_easy_backbones} indicate that the \asp{} attack no longer outperforms the \textit{Swap} baseline by a statistically significant margin.
\cref{fig:protonets_easy_backbones} considers three scenarios in increasing order of difficulty. For Scenario A, the perturbation size is very large ($\epsilon=1$) and the entire support set has been adversarially perturbed. At this scale, tampering could be clearly inferred by inspection and it is unlikely that an adversary would have control over the entire support set, so this is the weakest scenario. In this setting, we see good transfer between the surrogate model, Resnet18 A with FiLM, and the target models (Resnet18 A without FiLM and Resnet18 B without FiLM). Scenario B reduces the perturbation size to be mostly imperceptible ($\epsilon=0.3$), but still perturbs the entire support set. Here, we start to see a reduction in attack effectiveness when transferring to Resnset18 B. In Scenario C, we use $\epsilon=0.3$ and also reduce the proportion of poisoned samples in the support set to $50\%$. Here we see significant reduction in transfer, even to Resnet18 A without FiLM. We also note that unlike the white box attack, we do not perform significantly better than the \textit{Swap} attack, and possibly perform a little worse.
%
% From rebuttal: Relative drop in accuracy when transferring adversarial support attacks from \cnaps{}, which uses a ResNet18 neural network, to fine-tuners that use ResNet18 and MNASNet neural networks on \metadataset{}, averaged over 100 tasks. The attacks were generated with $\epsilon=0.05$, $\gamma=0.0015$, $L = 100$, with all classes but only $20\%$ of the shots adversarial.
%\begin{figure*}
	%\centering
	%\includegraphics[width=0.9\textwidth]{figures/few_shot_to_finetuner_resnet.pdf}
	%\caption{}
	%\label{fig:basic_meta_dataset}
%\end{figure*}
%
%\begin{figure*}
	%\centering
	%\includegraphics[width=0.9\textwidth]{figures/few_shot_to_finetuner_mnasnet.pdf}
	%\caption{}
	%\label{fig:basic_meta_dataset}
%\end{figure*}
%
% Also have heat map for small scale which shows similar thing
%
%\begin{figure*}
	%\centering
	%\includegraphics[width=0.9\textwidth]{figures/few_shot_to_finetuner_mnasnet.pdf}
	%\caption{}
	%\label{fig:basic_meta_dataset}
%\end{figure*}
%
We know from literature that adversarial attacks transfer well and should thus be fairly robust to the kind of feature extractor being used. Since we're only changing the feature extractor and keeping the type of learner the same, we expect the support attacks, which are tailored to fool the learning mechanism, to have a benefit over the swap attack. However, this is not borne out in experiments.
We consider two avenues of explanation that account for the poor transfer between feature extractors: mismatch between the target and surrogate models, and ``overfitting'' of the attack to the support set and/or surrogate model. We consider a number of mitigation strategies intended to address these two problems below:
% Sensitivity of FiLM parameters, surrogate models
%
% Regularisation by dropout and gaussian FiLM perturbations; context shuffle
%
% Will a surrogate model help?
\paragraph{Decision Boundary Alignment} Since the target and surrogate models have different feature extractors, it may be that their embedding spaces and consequently also their decision boundaries are very different. We can try to ``re-align'' the surrogate model's decision boundaries by training it to produce the same predictions as the target model, an approach which has been used in literature when perpetrating transfer attacks \citep{papernot2016practical}. In a meta-learning setting, this corresponds to relabeling the support set presented to the surrogate model based on the target model's predictions. The labels of the support set when presented to the surrogate model will thus not be the true labels, but rather the labels assigned to the support images by the target model. The relabeled support set is also used to generate the swap attack so that the comparison is fair.
\begin{figure*}
	\centering
	\includegraphics[width=1\textwidth]{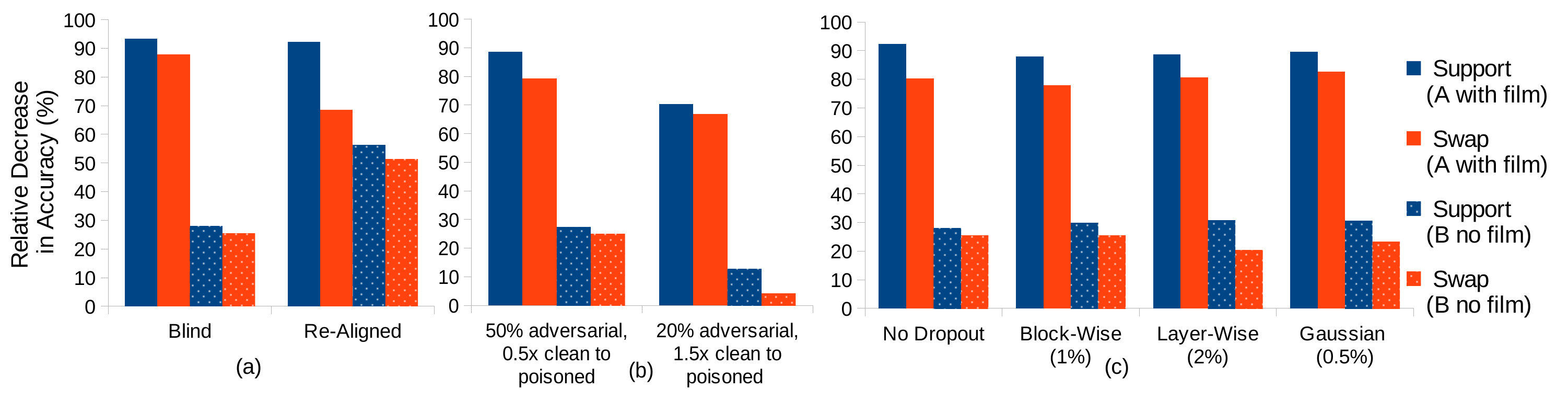}
	\caption{Mitigation strategies for transfer attacks on CIFAR-100. Solid fill indicates attacks on the surrogate (Resnset18 A with FiLM); dots indicate transfer attacks to Resnet18 B no FiLM. Support set is $50\%$ adversarial, $\epsilon=0.3$, $\gamma=0.0167$, $L=100$. (a) In \textit{Blind}, the surrogate has no knowledge of the target's decision boundaries; in \textit{Re-Aligned} the support set presented to the surrogate has been relabeled to match the target's predictions. (b) Clean support images are varied during attack generation. (c) Effect of different dropout strategies.}
	\label{fig:protonets_mitigations}
\end{figure*}
As shown in \cref{fig:protonets_mitigations}(a), relabeling the support set does significantly improve transfer compared to the ``blind'' scenario in which the surrogate model has no knowledge about the target model's decision boundaries. However, \asp{} is still not significantly better than the swap attack.
\paragraph{Hot Start} 
% Will a hot start help?
The \asp{} optimization procedure may be finding a different minimum, which is less robust to different feature extractors than that found by the swap attack. We propose to \textit{hot-start} the \asp{} attack, where given a support set $D_S$, we spend half the optimization budget performing a query attack to produce ${D}_S'$, which is then used as input to an adversarial support set attack to produce $\tilde{D}_S$ with the other half of the budget. In this way, we start the \asp{} attack off in the region of the local minimum found by the query attack.
However, as shown in \cref{fig:basic_meta_dataset}, the hot start does not improve on the normally initialized \asp{} attack and may be worse in some scenarios.
%
% Will shuffling the context set reduce overfitting?
\paragraph{Support Set Shuffling} 
Simple \cnaps{} makes use of a set encoder to produce a representation of the entire support set and so the attack must effectively manipulate the poisoned examples so that the embedded representation of the poisoned support set is sufficiently different to affect predictions. This may explain why we observe that the poisoning attack requires a high proportion of poisoned examples in the support set in order to be effective. 
The potency of the poisoned examples thus depend on the rest of the (unperturbed) support set. When transferred, the support set may embed very differently, especially if a large proportion of it is not adversarial,  negating the effects of the poisoned points. In order to generate solutions that are less dependent on the rest of the support set we propose a procedure whereby the clean elements in the support set are not kept fixed for the duration of attack generation, but instead shuffle different clean images in and out of the support set.
\cref{fig:protonets_mitigations}(b) considers two configurations: each varying the percentage of the support set that is poisoned (50\% and 20\%, respectively) and the proportion of clean images in the support set during attack generation (the clean : poisoned ratio is $0.5:1$ and $1.5:1$, respectively). Note that we are careful not to assume access to more data than any of the other scenarios. The first setting performs no better than an attack generated without shuffling.
The second setting performs worse -- although the attack has access to a more varied support set, this comes at the price of poisoning a smaller proportion of the support set since we cannot access additional data.
%
% Will regularisation by dropout/FiLM noise reduce overfitting?
\paragraph{Dropout} If the attack optimization procedure is overfitting to the surrogate model, then regularization by dropout is an established remedy. By applying dropout to the feature extractor, we may hope for the resulting attack to generalize better to new feature extractors.
%
%Since we may expect the FiLM parameters to be fair source of model mismatch (why?), we apply dropout to the FiLM parameters. 
%
We drop out the FiLM parameters of the surrogate's feature extractor using one of three different strategies: (1) Block-Wise - the FiLM parameters for an entire Resnet block is either on/off, i.e. either computed as usual or we set $\zeta=1.0$, $\beta = 0$; (2) Layer-Wise - each FiLM layer within a Resnet block may be switched on or off individually, i.e. compute $\zeta$ and $\beta$ as usual, but apply a dropout mask; (3) Gaussian Dropout - the FiLM parameters are smoothed by a Gaussian with standard deviation given by $\sqrt{{d}/(1-d)}$ where $d$ is the dropout probability.
Although \cref{fig:protonets_mitigations}(c) shows slight improvement in transferability, the difference between the \asp{} and swap attacks remain statistically insignificant. 
%
%
%
%) that amortised meta-learners let us jointly craft sets of datapoints that when used as training data can fool a classifier, (ii) jointly crafting a set of datapoints might be expected to synergistically fool a classifier allowing for very strong data-poisoning attacks that would be hard to detect.
\section{Conclusion}
\label{sec:conclusion}
%In this paper, we frame adversarial attacks in the context of meta-learners, both in terms of poisoning and evasion (i.e.~query) attacks. 
In this paper, we propose \asp{}, an attack against amortized meta-learners that jointly crafts sets of colluding points which can manipulate a meta-learner when used as training data.
We expect \asp{} attacks to be strong because the attack images are generated by backpropagating through the learner's adaptation process with respect to the loss over an entire set of test inputs and are thus able to collude to fool the classifier. We may also expect \asp{} enjoy the same robustness to different feature extractors as a standard evasion attack.
We show that although \asp{} is very effective in a white box setting, it does not transfer well, even to the same learner with a different feature extractor. 

We explore a number of strategies to improve the transferability of \asp{}, but in all cases \asp{} was not able to transfer significantly better than the swap attack baseline (which performs a query or evasion attack in the usual way and simply swaps it into the meta-learner's support set).
Even though the swap attack should be much weaker, since it is generated to fool the learner only for the image in question, it proves a very effective baseline for poisoning few-shot learners. 
Since an attacker will rarely have access to the internal workings of the target model, the transferability of an attack is an important consideration, which prevents the deployment of \asp{} in real-life. But future work may glean insight into making models more robust.
%. 
%Since the \asp{} attack backpropagates through the learner's adaptation process and should at first glance enjoy the same robustness to different feature extractors as a standard evasion attack, we expect an \asp{} attack to be more effective than the swap attack.
%
%However, swap attacks -- where a simple adversarial example generated in the usual way is inserted as poison into the support set -- proves a very effective baseline. 
%
%Where query attacks generate a single adversarial image at test-time, we generate an adversarial set of inputs that are generated to affect the classifier's performance on a set of test images.
%
%
%The query attack image is generated to fool the learner only for the image in question and backpropagation is only through the adapted learner, not the learning process
%er than simply inserting a query attack into the dataset 
% can collude to fool the classifier and 
%n adversarial attack generated in the usual  generating a set of colluding inputs that, when given the learner as input, significantly reduces test-time accuracy. 
%We expect the colluding set or \asp{} propose an attack that generates colluding sets of inputs that can be given to a meta-learner as inpu t
%
%. Even though adversarial attacks are known to transfer well in literature, we find that adversarial support attacks do not transfer

\begin{ack}
This work was performed using resources provided by the Cambridge Service for Data Driven Discovery (CSD3) operated by the University of Cambridge Research Computing Service (\url{www.csd3.cam.ac.uk}), provided by Dell EMC and Intel using Tier-2 funding from the Engineering and Physical Sciences Research Council (capital grant EP/T022159/1), and DiRAC funding from the Science and Technology Facilities Council (\url{www.dirac.ac.uk}).

Richard E. Turner is supported by funding from Google, Amazon, ARM, Improbable and Microsoft. The paper built on software developed under Prosperity Partnership EP/T005386/1 between Microsoft Research and the University of Cambridge.
\end{ack}

\bibliography{references.bib}
\bibliographystyle{plainnat}

\clearpage

\appendix

\section{Appendix}

\subsection{Episodic Training}
\label{app:episodic_training}
\begin{figure}[b]
	\centering
	\includegraphics[width=0.9\linewidth]{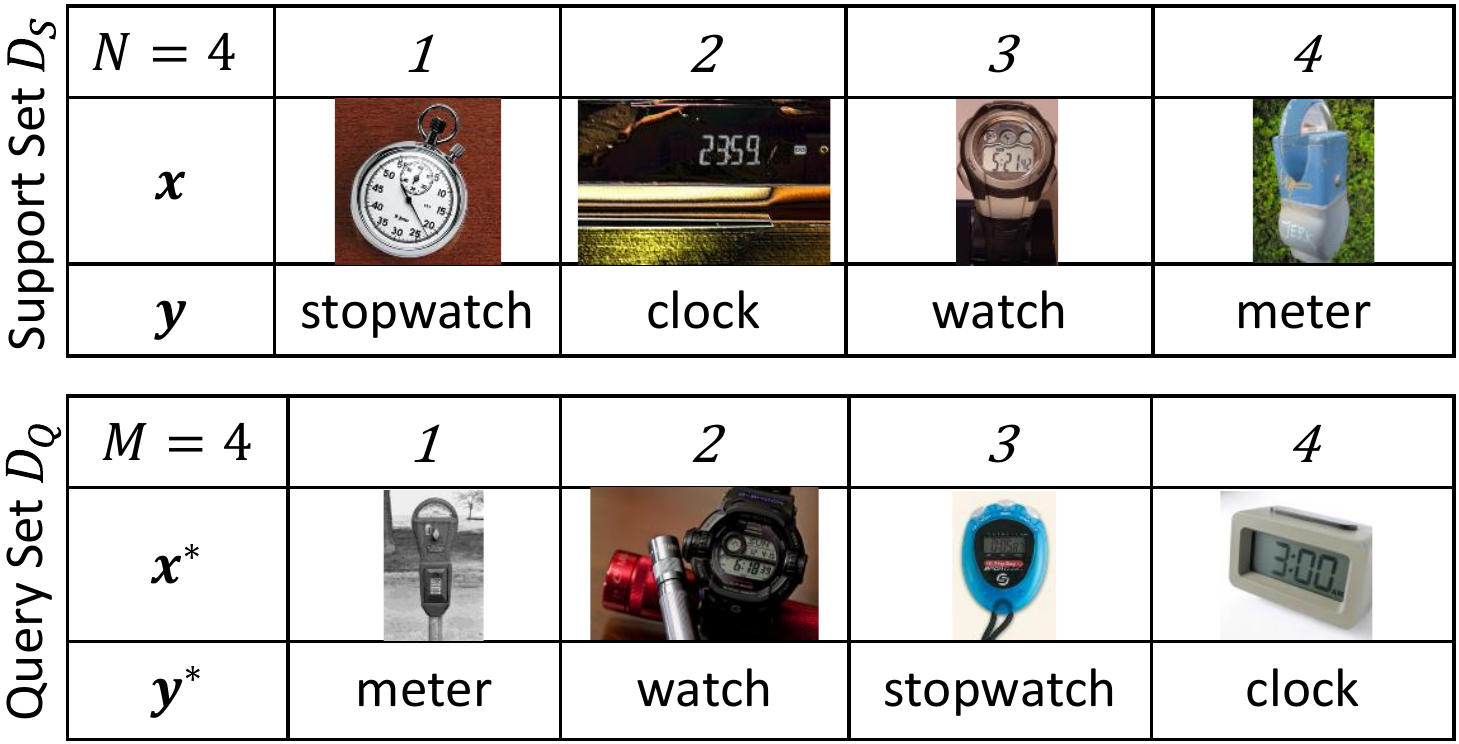}
	\caption{Example task with $C=4$ classes with $N=M=4$.}
	\label{fig:task}
\end{figure}

There has been an explosion of meta-learning based few-shot learning algorithms proposed in recent years. For an in-depth review see \citet{hospedales2020meta}. The majority of modern meta-learning methods employ \textit{episodic} training \citep{vinyals2016matching}.
During meta-training, a task $\tau$ is drawn from $p(\tau)$ and randomly split into a support set $D_S$ and query set $D_Q$. \cref{fig:task} depicts an example few-shot classification task.
The meta-learner $g$ takes as input the support set $D_S$ and produces task-specific classifier parameters $\vpsi = g(D_S)$ which are used to adapt the classifier $f$ to the current task. The classifier can now make task-specific predictions $f(x^{\ast}, \vpsi= g(D_S))$ for any test input $x^{\ast} \in D_Q$. Refer to \textit{Clean} in \cref{fig:attacks}.
A loss function $\mathcal{L}(f(x^{\ast}, \vpsi), y^{\ast})$ then computes the loss between the predictions for the label $f(x^{\ast}, \vpsi)$ and the true label $y^{\ast}$.
Assuming that $\mathcal{L}$, $f$, and $g$ are differentiable, the meta-learning algorithm can then be trained with stochastic gradient descent by back-propagating the loss and updating the parameters of $f$ and $g$.

\subsection{Few-shot Learner Meta-training Protocols}
\label{app:training_protocols}
In the following, the meta-training protocols for the few-shot learners used in the experiments.

\subsubsection{Datasets}

\paragraph{\miniimagenet{}}
\miniimagenet{} is a subset of the larger Imagenet dataset \citep{russakovsky2015imagenet} created by \citet{vinyals2016matching}. It consists of 60,000 color images that is sub-divided into 100 classes, each with 600 instances. The images have dimensions of $84\times 84$ pixels. \citet{ravi2016optimization} standardized the 64 training, 16 validation, and 20 test class splits. \miniimagenet{} has become a defacto standard dataset for benchmarking few-shot image classification methods with the following classification task configurations:
\begin{inlinelist}
\item 5-way, 1-shot;
\item 5-way, 5-shot.
\end{inlinelist}

\paragraph{\metadataset{}}
\label{app:meta_dataset}
\metadataset{} \citep{triantafillou2019meta} is composed of ten (eight train, two test) image classification datasets. We augment Meta-Dataset with three additional held-out datasets: MNIST \citep{lecun2010mnist}, CIFAR10 \citep{krizhevsky2009learning}, and CIFAR100 \citep{krizhevsky2009learning}. The challenge constructs few-shot learning tasks by drawing from the following distribution. First, one of the datasets is sampled uniformly; second, the ``way'' and ``shot'' are sampled randomly according to a fixed procedure; third, the classes and support / query instances are sampled. Where a hierarchical structure exists in the data (ImageNet or Omniglot), task-sampling respects the hierarchy. In the meta-test phase, the identity of the original dataset is not revealed and the tasks must be treated independently (i.e.~no information can be transferred between them). Notably, the meta-training set comprises a disjoint and dissimilar set of classes from those used for meta-test. \metadataset{} is presently, the "gold standard" for evaluating few-shot classification methods. Full details are available in \citet{triantafillou2019meta}.

In our experiments, we excluded the Omniglot,  Textures, Fungi, and Traffic Signs datasets from evaluation because their test splits are too small to allow for a fair assessment of the attack's generalization, even though the attacks reduced the classification accuracy on those datasets to approximately zero in the \textit{\asp{} Specific} case.

For \metadataset{}, we meta-trained Simple \cnaps{} using the code from \citet{requeima2019code} with FiLM feature adaptation. We made modifications to the code to enable various adversarial attacks. The meta-trained model attained the following results:

ilsvrc 2012: $55.1\pm1.1$,
omniglot: $90.8\pm0.6$,
aircraft: $82.3\pm0.6$,
cu birds: $74.0\pm0.9$,
dtd: $63.4\pm0.7$,
quickdraw: $75.3\pm0.8$,
fungi: $44.6\pm1.0$,
vgg flower: $90.3\pm0.5$,
traffic sign: $67.9\pm0.8$,
mscoco: $40.8\pm1.0$,
mnist: $91.4\pm0.5$,
cifar10: $72.5\pm0.8$,
cifar100: $58.4\pm1.0$.

\subsection{Attack Summary}
\label{app:attack_summary}
\begin{figure*}
	\centering
	\includegraphics[width=1.0\textwidth]{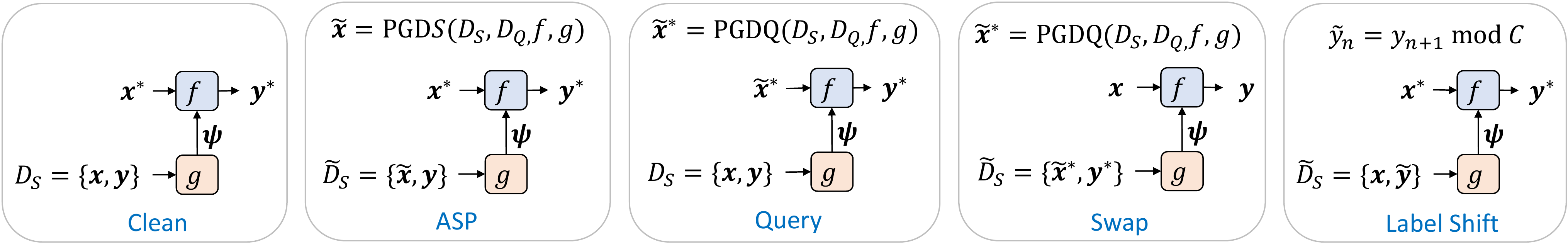}
	\caption{Attacks on meta-learning based few-shot image classifiers. $f$ and $g$ denote the classifier and trained meta-learner, respectively. Each diagram depicts how an attack is applied and includes an expression for the attack's computation using Algorithms \ref{alg:pgd} and \ref{alg:pgd_query}.}
	\label{fig:attacks}
\end{figure*}

We summarize the types of attacks we perform in \cref{fig:attacks}. The first scenario, \textit{Clean}, illustrates how the meta-learner $g$ performs a test-time task, taking the support set $D_S$ as input to produce 
 parameters $\vpsi = g(D_S)$ which are used to adapt the classifier $f$ to the task. The classifier makes task-specific predictions $f(x^{\ast}, \vpsi= g(D_S))$ for any test input $x^{\ast} \in D_Q$. \textit{ASP} and \textit{Query} illustrate the \asp{} and Query attacks as discussed in \cref{sec:attacking_learners}. \textit{Swap} illustrates a swap attack, which is used as a baseline comparison for \asp{},  where a set of images are perturbed with a query attack and then inserted into the support set. Query attacks are typically cheaper to compute, since they do not require back-propagation through the meta-learner, so it is an important baseline to consider.
%\textit{Label Shift} is a simple attack on the support set which involves mislabelling the support set images by shifting the true label index by one in a modulo arithmetic fashion. We consider systematic mislabeling in this way to be a strong attack for comparison, though this ``attack'' would be easily detected by inspection. We thus only consider the label shift baseline for the small scale results presented in \cref{app:small_scale}.

\subsection{Additional Experimental Details}
\label{app:details}

 In our experiments, all the input images were re-scaled to have pixel values between $-1$ and $1$. We considered perturbations using the $\ell_\infty$ norm, on a scale of $[-1, 1]$, so that $\epsilon = 0.1$ corresponds to allowing $\pm10\%$ or an absolute change of $\pm0.2$ to the intensity of each pixel in an image.

We calculated the perturbation step size $\gamma$ to depend on $\epsilon$ and the maximum number of iterations, so that $\gamma = r \frac{\epsilon}{L}$, where $r$ is a scaling coefficient.

\subsection{Unnormalised Large-Scale Attack Results}
\label{app:large_scale_results}

This section of the appendix presents the unnormalized results for all figures, along with 95\% confidence intervals.

In \cref{tab:protonets_on_metadataset} we present the unnormalized numbers for \cref{fig:basic_meta_dataset}, which shows the efficacy of \asp{} in a white box setting. \cref{tab:protonets_backbone} considers transfer between a surrogate model and target models with different feature extractors as presented in \cref{fig:protonets_easy_backbones}\textit{(Left)}. Tables \ref{tab:hot_start_0.3} and \ref{tab:hot_start_0.1} consider the effect of the hot-start mitigation strategy for different perturbation sizes as presented in \cref{fig:protonets_easy_backbones}\textit{(Right)}. Tables \ref{tab:realign}, \ref{tab:shuffle_context} and \ref{tab:dropout} correspond to \cref{fig:protonets_mitigations} and show the effect of other mitigation strategies, specifically decision boundary re-alignment, shuffling different clean inputs into the support set, and dropout. All numbers are percentages and the $\pm$ sign indicates the 95\% confidence interval.

\begin{table}\centering
\caption{Accuracy of Simple \cnaps{} on the \metadataset{} benchmark in the \textit{Clean}, \textit{Specific} and \textit{General} scenarios when attacking with an adversarial support set, with $\epsilon=0.05$, $\gamma=0.0015$, $L = 100$, with all classes, but only $20\%$ of the shots poisoned. All figures are percentages and the $\pm$ sign indicates the 95\% confidence interval over 500 tasks.}\label{tab:protonets_on_metadataset}
\vskip 0.15in
\begin{adjustbox}{max width=\columnwidth}
\begin{tabular}{lccccc}\toprule
&Clean &Specific &General &Swap \\\midrule
ilsvrc\_2012 &52.2$\pm$0.2 &1.5$\pm$0.1 &14.8$\pm$0.1 &42.8$\pm$0.2 \\
aircraft &78.5$\pm$0.5 &0.0$\pm$0.0 &7.2$\pm$0.2 &65.3$\pm$1.0 \\
cu\_birds &71.4$\pm$1.1 &1.8$\pm$0.2 &8.9$\pm$0.4 &50.6$\pm$2.0 \\
quickdraw &74.1$\pm$0.1 &32.8$\pm$1.2 &41.3$\pm$0.2 &57.7$\pm$0.3 \\
vgg\_flower &89.1$\pm$0.6 &10.6$\pm$0.8 &27.7$\pm$1.0 &78.7$\pm$1.8 \\
traffic\_sign &35.2$\pm$0.4 &4.1$\pm$0.3 &12.3$\pm$0.3 &24.8$\pm$0.5 \\
mscoco &44.1$\pm$0.3 &2.9$\pm$0.1 &10.8$\pm$0.1 &29.2$\pm$0.3 \\
mnist &90.3$\pm$0.1 &58.8$\pm$1.4 &68.9$\pm$0.2 &85.8$\pm$0.2 \\
cifar10 &64.6$\pm$0.1 &1.6$\pm$0.1 &13.5$\pm$0.1 &51.8$\pm$0.2 \\
cifar100 &53.5$\pm$0.6 &2.4$\pm$0.1 &7.7$\pm$0.2 &25.0$\pm$0.8 \\
\bottomrule
\end{tabular}
\end{adjustbox}
\end{table}

\begin{table}[!ht]
\caption{Transfer from the surrogate model (Simple \cnaps{} using Resnet A with FiLM) to Simple \cnaps{} using alternative feature extractors on CIFAR-100. Scenario A: $\epsilon = 1.0$ with entire support set adversarial; Scenario B: $\epsilon=0.3$ with entire support set adversarial; Scenario C: $\epsilon=0.3$ with 50\% of support set adversarial. Attacks are generated using PGD with $L=100$. These results correspond to \cref{fig:protonets_easy_backbones} (\textit{Left}).}
\label{tab:protonets_backbone}
    \centering
    \begin{adjustbox}{max width=\columnwidth}
    \begin{tabular}{ccccc}
    \toprule
        ~ & ~ & Clean & ASP & Swap  \\ 
        \cmidrule{2-5}
        Surrogate  & Scenario A & 52.4$\pm$2.1 & 3.6$\pm$0.4 & 2.9$\pm$0.4  \\ 
        (Resnet A FiLM) & Scenario B & 54.2$\pm$1.2 & 3.2$\pm$0.2 & 4.9$\pm$0.3  \\ 
        ~ & Scenario C & 54.5$\pm$1.8 & 3.6$\pm$0.3 & 6.6$\pm$0.6  \\ 
        \cmidrule{2-5}
        Transfer to & Scenario A & 48.7$\pm$2.1 & 4.7$\pm$0.5 & 3.9$\pm$0.5  \\ 
        Resnet A (no FiLM) & Scenario B & 50.2$\pm$1.2 & 4.6$\pm$0.3 & 4.0$\pm$0.3  \\ 
        ~ & Scenario C & 50.5$\pm$3.4 & 32.7$\pm$2.0 & 28.2$\pm$1.3  \\ 
        \cmidrule{2-5}
        Transfer to & Scenario A & 38.6$\pm$2.0 & 4.9$\pm$0.6 & 4.6$\pm$0.6  \\ 
        Resnet B (no FiLM) & Scenario B & 39.1$\pm$1.2 & 10.2$\pm$0.6 & 9.0$\pm$0.5  \\ 
        ~ & Scenario C & 40.1$\pm$1.9 & 28.9$\pm$2.0 & 29.9$\pm$1.6 \\ 
    \bottomrule
    \end{tabular}
\end{adjustbox}
\end{table}

\begin{table}[!ht]
\caption{Effect of hot-starting the ASP attack on transferability from the surrogate model (Simple \cnaps{} using Resnet A without FiLM) to Simple \cnaps{} using an alternative feature extractor on ILSVRC 2012. Attacks were generated using PGD with  $\epsilon=0.3$ and $L=100$. These results correspond to \cref{fig:protonets_easy_backbones} (\textit{Right}) for $\epsilon=0.3$. }
\label{tab:hot_start_0.3}
    \centering
    \begin{adjustbox}{max width=\columnwidth}
    \begin{tabular}{lcccc}
    \toprule
        ~ & Clean & ASP & Swap & Hot Start  \\ 
        \midrule
        Surrogate (Resnet18 A no FiLM) & 52.1$\pm$0.7 & 20.1$\pm$0.4 & 46.3$\pm$0.9 & 22.0$\pm$0.4  \\ 
        Transfer to Resnet18 A (with FiLM) & 55.8$\pm$0.7 & 31.5$\pm$0.7 & 28.7$\pm$0.5 & 35.2$\pm$0.7  \\ 
        Transfer to Resnet18 B (no FiLM) & 43.6$\pm$0.7 & 36.1$\pm$0.7 & 32.2$\pm$0.6 & 36.4$\pm$0.7 \\ 
    \bottomrule
    \end{tabular}
\end{adjustbox}
\end{table}

\begin{table}[!ht]
\caption{Effect of hot-starting the ASP attack on transferability from the surrogate model (Simple \cnaps{} using Resnet A without FiLM) to Simple \cnaps{} using an alternative feature extractor on ILSVRC 2012. Attacks were generated using PGD with  $\epsilon=0.1$ and $L=100$. These results correspond to \cref{fig:protonets_easy_backbones} (\textit{Right}) for $\epsilon=0.1$. }
\label{tab:hot_start_0.1}
    \centering
    \begin{adjustbox}{max width=\columnwidth}
    \begin{tabular}{lcccc}
    \toprule
        ~ & Clean & ASP & Swap & Hot Start  \\ 
        \midrule
        Surrogate (Resnet18 A no FiLM) & 54.1$\pm$0.8 & 16.0$\pm$0.4 & 50.9$\pm$0.8 & 16.1$\pm$0.4  \\ 
        Transfer to Resnet18 A (with FiLM) & 58.1$\pm$0.8 & 46.2$\pm$0.8 & 41.1$\pm$0.7 & 48.5$\pm$0.8  \\ 
        Transfer to Resnet18 B (no FiLM) & 45.7$\pm$0.7 & 41.8$\pm$0.7 & 35.7$\pm$0.6 & 43.1$\pm$0.7 \\ 
    \bottomrule
    \end{tabular}
\end{adjustbox}
\end{table}

\begin{table}[!ht]
\caption{Effect of decision boundary realignment when transferring attacks from the surrogate (Simple \cnaps{} using Resnset18 A with FiLM) to the target (Simple \cnaps{} using Resnet18 B no FiLM). The support set is $50\%$ adversarial. Attacks are generated using PGD with $\epsilon=0.3$, $\gamma=0.0167$, $L=100$. In \textit{Blind}, the surrogate has no knowledge of the target's decision boundaries; in \textit{Re-Aligned} the support set presented to the surrogate has been relabeled to match the target's predictions. This corresponds to \cref{fig:protonets_mitigations}(a). }
\label{tab:realign}
    \centering
    \begin{adjustbox}{max width=\columnwidth}
    \begin{tabular}{ccccc}
    \toprule
        ~ & ~ & Clean & ASP & Swap  \\ 
        \midrule
        Surrogate & Re-Aligned & 52.6$\pm$2.0 & 4.0$\pm$0.4 & 16.5$\pm$1.9  \\ 
         (Resnet18 A with FiLM) & Blind & 54.5$\pm$1.8 & 3.6$\pm$0.3 & 6.6$\pm$0.6  \\
        \cmidrule{2-5}
        Transfer to  & Re-Aligned & 43.9$\pm$2.1 & 19.2$\pm$1.2 & 21.4$\pm$0.9  \\ 
        Resnet18 B (no FiLM) & Blind & 40.1$\pm$1.9 & 28.9$\pm$2.0 & 29.9$\pm$1.6 \\
    \bottomrule
    \end{tabular}
\end{adjustbox}
\end{table}

\begin{table}[!ht]
    \centering
    \caption{Effect of varying the clean images in the support set when transferring attacks from the surrogate (Simple \cnaps{} using Resnset18 A with FiLM) to the target (Simple \cnaps{} using Resnet18 B no FiLM). Attacks are generated using PGD with $\epsilon=0.3$, $\gamma=0.0167$, $L=100$. For Setting A, 50\% of the support set is adversarial and the ratio of clean to poisoned patterns is 0.5:1. For Setting B, 20\% of the support set is poisoned and the ratio of clean to poisoned patterns is 1.5:1. This table corresponds to \cref{fig:protonets_mitigations}(b). }
\label{tab:shuffle_context}
    \begin{adjustbox}{max width=\columnwidth}
    \begin{tabular}{ccccc}
    \toprule
        ~ & ~ & Clean & ASP & Swap  \\ 
        \midrule
        Surrogate & Setting A & 55.7$\pm$2.2 & 6.3$\pm$0.6 & 11.5$\pm$1.3  \\ 
        (Resnet18 A with FiLM)  & Setting B & 68.8$\pm$0.4 & 32.4$\pm$0.4 & 44.6$\pm$0.7  \\ 
        \cmidrule{2-5}
        Transfer to  & Setting A  & 41.3$\pm$2.1 & 30.0$\pm$2.1 & 31.0$\pm$1.5  \\ 
        Resnet18 B (no FiLM) & Setting B & 55.1$\pm$0.4 & 51.7$\pm$0.5 & 48.4$\pm$0.3 \\ 
    \bottomrule
    \end{tabular}
\end{adjustbox}
\end{table}

\begin{table}[!ht]
\caption{Effect of various dropout strategies when transferring attacks from the surrogate (Simple \cnaps{} using Resnset18 A with FiLM) to the target (Simple \cnaps{} using Resnet18 B no FiLM). The support set is $50\%$ adversarial. Attacks are generated using PGD with $\epsilon=0.3$, $\gamma=0.0167$, $L=100$. This table corresponds to \cref{fig:protonets_mitigations}(c). }
\label{tab:dropout}
    \centering
    \begin{adjustbox}{max width=\columnwidth}
    \begin{tabular}{ccccccc}
    \toprule
        ~ & \multicolumn{3}{c}{Surrogate (Resnet18 A with FiLM)}  & \multicolumn{3}{c}{Transfer to Resnet18 B (no FiLM)}   \\
        \midrule
        ~ & Clean & ASP & Swap & Clean & ASP & Swap  \\ 
        \cmidrule(l{2pt}r{2pt}){2-4} 
        \cmidrule(l{2pt}r{2pt}){5-7}
        No dropout & 55.5$\pm$1.4 & 4.2$\pm$0.3 & 10.9$\pm$1.1 & 40.1$\pm$1.9 & 28.9$\pm$2.0 & 29.9$\pm$1.6  \\ 
        Block-Wise (1\%) & 44.6$\pm$1.5 & 5.4$\pm$0.3 & 9.8$\pm$0.7 & 42.3$\pm$1.8 & 29.7$\pm$1.8 & 31.5$\pm$1.3  \\ 
        Layer-Wise (2\%) & 44.5$\pm$2.1 & 5.0$\pm$0.5 & 8.6$\pm$1.0 & 38.3$\pm$1.9 & 26.5$\pm$1.8 & 30.5$\pm$1.5  \\ 
        Gaussian (0.5\%) & 47.4$\pm$1.6 & 4.9$\pm$0.4 & 8.2$\pm$1.0 & 39.0$\pm$1.5 & 27.1$\pm$1.4 & 29.9$\pm$1.2 \\ 
    \bottomrule
    \end{tabular}
\end{adjustbox}
\end{table}

\clearpage

\subsection{Query Attacks}
\label{app:pgd_query_algorithm}
We present our algorithm for performing query attacks with PGD in \cref{alg:pgd_query}.
\begin{algorithm}[h]
\caption{PGD for Query Attack}
\label{alg:pgd_query}
\begin{algorithmic}[1]
\Require
\Statex $I_{min}$: Minimum image intensity
\Statex $I_{max}$: Maximum image intensity
\Statex $L$: Number of iterations
\Statex $\epsilon$: Perturbation amount
\Statex $\gamma$: Step size
\Statex $D_S \equiv \{\vx,\vy\}$
\Statex $D_Q \equiv \{\vx^*,\vy^*\}$
%\LineComment{We use cross-entropy loss for $\mathcal{L}$.}
\item[]
\Procedure{PGDQ}{$D_S,D_Q,f,g$}
\State $\vdelta \sim U(-\epsilon, \epsilon)$
\State $\tilde \vx^* \leftarrow \text{clip}(\vx^* + \vdelta, I_{min}, I_{max})$ 
\For{$n \in 1,...,L$}
\State $\vdelta \leftarrow \text{sgn}(\nabla_{\tilde \vx^*}\mathcal{L}(f(\tilde x^*, g(x, y)),y^{\ast})$
\State $\tilde \vx^* \leftarrow \text{clip}(\tilde \vx^* + \gamma\vdelta, I_{min}, I_{max})$
\State $\tilde \vx^* \leftarrow x + \text{clip}(\tilde \vx^* - \vx^*, -\epsilon, \epsilon)$
\EndFor
\State \Return $\tilde \vx^*$
\EndProcedure
\end{algorithmic}
\end{algorithm}

\end{document}